\documentclass[letterpaper]{article} % DO NOT CHANGE THIS
\usepackage{aaai2026}  % DO NOT CHANGE THIS
\usepackage{times}  % DO NOT CHANGE THIS
\usepackage{helvet}  % DO NOT CHANGE THIS
\usepackage{courier}  % DO NOT CHANGE THIS
\usepackage[hyphens]{url}  % DO NOT CHANGE THIS
\usepackage{graphicx} % DO NOT CHANGE THIS
\usepackage{natbib}  % DO NOT CHANGE THIS AND DO NOT ADD ANY OPTIONS TO IT
\usepackage{caption} % DO NOT CHANGE THIS AND DO NOT ADD ANY OPTIONS TO IT
\usepackage{algorithm}
\usepackage{algorithmic}
\usepackage{amsmath}
\usepackage{enumitem}
\usepackage{booktabs}
\usepackage{newfloat}
\usepackage{listings}
\DeclareCaptionStyle{ruled}{labelfont=normalfont,labelsep=colon,strut=off} % DO NOT CHANGE THIS
\floatstyle{ruled}
\newfloat{listing}{tb}{lst}{}
\floatname{listing}{Listing}
\title{OTI: A Model-Free and Visually Interpretable Measure of Image Attackability}
\author {
    Jiaming Liang\textsuperscript{\rm 1},
    Haowei Liu\textsuperscript{\rm 2},
    Chi-Man Pun\textsuperscript{\rm 1}\footnote{Corresponding author.}
}
\affiliations {
    \textsuperscript{\rm 1}Faculty of Science and Technology, University of Macau, Macau, China\\
    \textsuperscript{\rm 2}Chongqing Key Laboratory of Image Cognition, Chongqing University of Posts and Telecommunications, China\\
    chinaliangjm@gmail.com, S230201069@stu.cqupt.edu.cn, cmpun@um.edu.mo
}
\begin{document}

\maketitle

\begin{abstract}
Despite the tremendous success of neural networks, benign images can be corrupted by adversarial perturbations to deceive these models. Intriguingly, images differ in their attackability. Specifically, given an attack configuration, some images are easily corrupted, whereas others are more resistant. Evaluating image attackability has important applications in active learning, adversarial training, and attack enhancement. This prompts a growing interest in developing attackability measures. However, existing methods are scarce and suffer from two major limitations: (1) They rely on a model proxy to provide prior knowledge (e.g., gradients or minimal perturbation) to extract model-dependent image features. Unfortunately, in practice, many task-specific models are not readily accessible. (2) Extracted features characterizing image attackability lack visual interpretability, obscuring their direct relationship with the images. To address these, we propose a novel \textbf{Object Texture Intensity (OTI)}, a model-free and visually interpretable measure of image attackability, which measures image attackability as the texture intensity of the image's semantic object. Theoretically, we describe the principles of OTI from the perspectives of decision boundaries as well as the mid- and high-frequency characteristics of adversarial perturbations. Comprehensive experiments demonstrate that OTI is effective and computationally efficient. In addition, our OTI provides the adversarial machine learning community with a visual understanding of attackability.
\end{abstract}

\begin{links}
    \link{Code}{https://github.com/chinaliangjiaming/OTI}
\end{links}

\section{Introduction}

Deep neural networks (DNNs) are playing a growing role in critical sectors, yet carefully crafted adversarial perturbations~\cite{szegedy2013intriguing} by adversaries can easily fool them, which incur significant threats. Hence, robustness analysis has long been a central theme in adversarial machine learning.

\begin{figure}[t]
    \centering
    \includegraphics[width=0.95\linewidth]{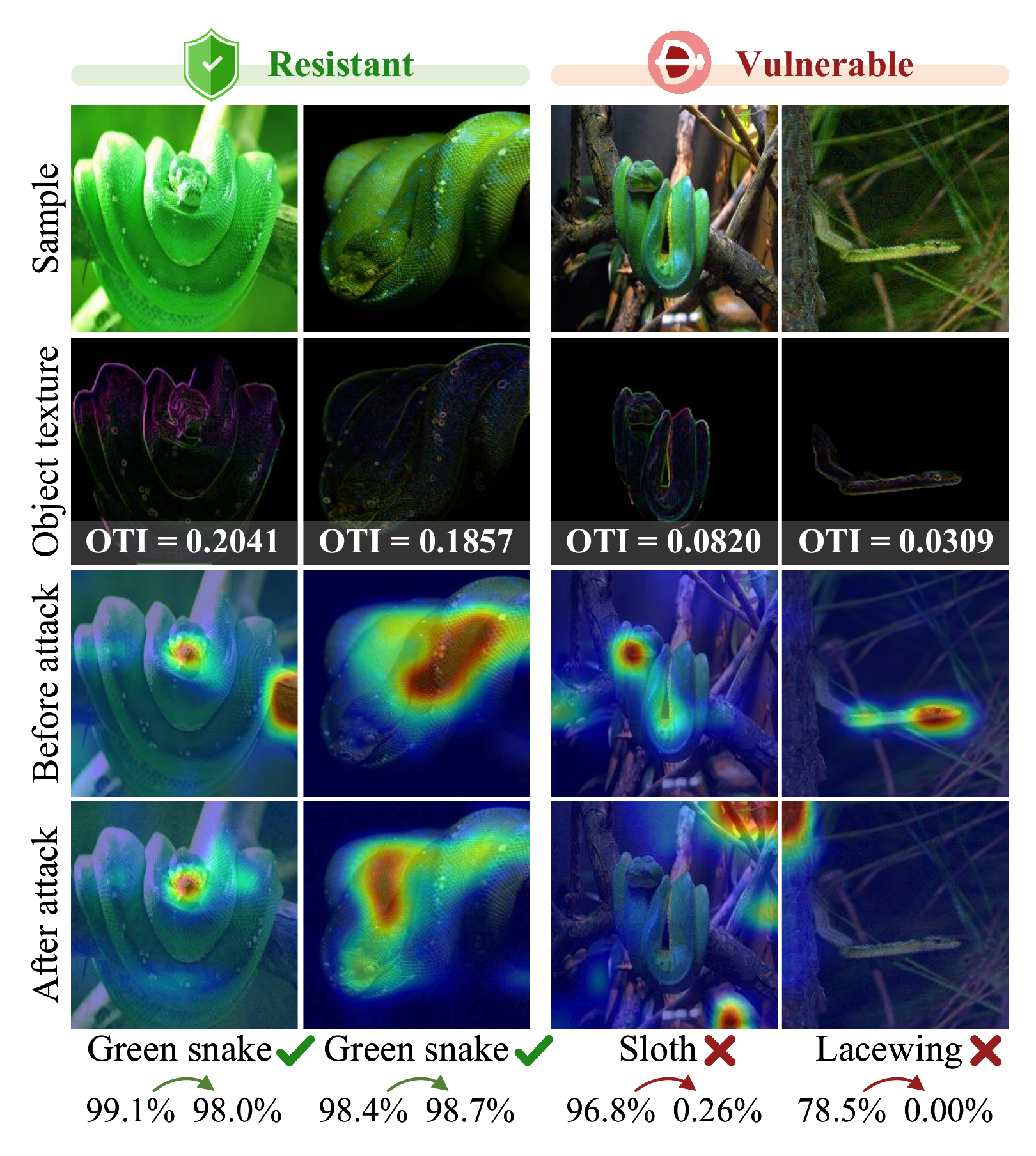}
    \caption{Measuring image attackability by object texture intensity (values in the second row). The third and fourth rows are Grad-CAM visualizations of benign images and adversarial examples under BSR attack (surrogate: R50, target: D161). The last two rows show the predicted labels of adversarial examples and the predicted probabilities of the true class for both benign images and adversarial examples.}
    \label{fig:introduction}
\end{figure}

Robustness analysis~\cite{liu2025comprehensive} from a \textit{model-centric} perspective has evolved over many years, spanning formal robustness certification~\cite{zhang2024seev}, robustness evaluation~\cite{yan2024enhance}, robust architecture design~\cite{xu2025rethinking}, and adversarial defense~\cite{liu2025comprehensive}. In contrast, recent work~\cite{raina2023identifying} proposes a novel robustness concept called image attackability from a \textit{sample-centric} viewpoint. Unlike the traditional focus on evaluating neural network robustness, this idea aims to assess the robustness\footnote{Higher robustness implies lower attackability and lower vulnerability. These three terms are interchangeable in this paper.} of \textit{benign images} against adversarial perturbations. In defense applications, image attackability evaluation helps identify the most informative images for active learning, select the most vulnerable images for efficient generation of adversarial examples for adversarial training~\cite{madry2017towards}, and construct effective subsets for debiasing~\cite{zhu2025evading}. 
% In addition, it can select the most vulnerable samples, thereby enabling more efficient generation of adversarial examples for adversarial training.
In attack applications, attackability evaluation enables more effective attacks by selecting the most vulnerable images. Furthermore, understanding image attackability can deepen our insight into the effectiveness of adversarial perturbations and thus advance the adversarial machine learning community.

Unfortunately, existing image attackability measures are scarce and face two major limitations: (1) \textit{Model-dependent}. IAARS~\cite{raina2023identifying} relies on a model proxy to train an image attackability discriminator and leverages its generalization capability to assess the attackability of unseen images on unseen models. ZGP~\cite{karunanayake2025quantifying} measures image attackability as the proportion of elements in the gradient of the proxy's loss with respect to the input whose absolute values are below a specified threshold. Both methods require a trained model proxy, which is often infeasible in many practical scenarios, especially where public models are not readily available (e.g., in the field of medical imaging) and the cost of training models is prohibitively high. Ideally, image attackability measures should work based only on the intrinsic attributes of the image itself and the task. (2) \textit{Visually uninterpretable.} Features characterizing image attackability extracted by existing methods lack visual interpretability for humans. IAARS encodes these features into the neural-network discriminator, and ZGP utilizes gradients. It is difficult to establish a visual connection between the extracted features and the image itself. Therefore, we aim to develop a \textit{model-free} and visually interpretable image attackability measure. 

Empirical results on visual tasks suggest that the smaller the semantic object~\cite{fan2021concealed} and the weaker the texture intensity, the more attackable the image. We explain these phenomena from the perspective of decision boundary~\cite{yan2024enhance} and the mid- and high-frequency nature~\cite{wang2020high} of adversarial perturbations. From the perspective of decision boundary, smaller semantic objects and weaker textures make images difficult to classify. As a result, their degradation vectors are greater and these images are closer from the decision boundary, leading to greater attackability. From the perspective of frequency domain, since adversarial perturbations are generally believed to rely heavily on mid- and high-frequency components, stronger semantic textures safeguard images against the adverse impact of the mid- and high-frequency components of adversarial perturbations and thus contribute to improved robustness.

Accordingly, we propose a novel \textbf{Object Texture Intensity (OTI)} measure, which quantifies image attackability by the sum of absolute texture values within the semantic object region, as shown in Figure~\ref{fig:introduction}. A lower OTI suggests a smaller semantic object and weaker texture intensity, indicating greater attackability. Extensive experiments demonstrate that OTI can effectively distinguish the image attackability across various advanced attacks, diverse domains, different tasks, varying dataset sizes, in both defense and non-defense scenarios, and across different configurations. The main contributions are summarized as follows:

\begin{itemize}
    \item To the best of our knowledge, this is the first work to reveal the relationship between the texture intensity of semantic objects and image attackability.

    \item We describe the relationship from the perspectives of decision boundary theory as well as the mid- and high-frequency nature of perturbations.
    
    \item Based on this relationship, we propose a novel image attackability measure, Object Texture Intensity (OTI). This is the first model-free and visually interpretable measure.

    % \item This paper establishes the theoretical foundation of OTI based on the high-dimensional decision boundary and the mid and high-frequency nature of perturbations.

    \item Our OTI provides a visual reference for human understanding of image attackability. Extensive experiments demonstrate that although OTI is model-free and simple, it is highly effective across different tasks, various attacks, and diverse configurations.
\end{itemize}

\section{Related Work}
In this section, we provide a comprehensive review of adversarial attacks and robustness analysis.

\subsection{Adversarial Attacks}
Adversarial attacks~\cite{szegedy2013intriguing} refer to the process of applying imperceptible perturbations to benign samples to induce decision deviations in the model. In the current landscape, according to the source of perturbations, adversarial attacks are generally classified into three major categories: gradient-based, query-based, and generation-based attacks.

\textit{Gradient-based attacks} exploit the input gradients of the loss to carry out gradient ascent, thus disrupting the model's decision. In the white-box settings, access to the full gradient information of the model enables nearly perfect attacks~\cite{kurakin2018adversarial, dong2018boosting}. In the black-box settings, adversaries typically exploit the adversarial transferability, whereby adversarial examples effective on one model are likely to succeed on other models. To improve transferability, transformation-based strategies~\cite{lin2024boosting, liang2025ic}, model-related strategies~\cite{ma2024improving}, ensemble-based strategies~\cite{chen2024rethinking, tang2024ensemble}, gradient-optimization strategies~\cite{wang2024boostinggi}, and objective-optimization strategies~\cite{zhou2018transferable, li2023improving} are employed.

\textit{Query-based attacks} rely solely on the outputs of the target model and iteratively update the perturbation until the attack succeeds or the query budget is exhausted. According to the type of output returned by the target model, query-based attacks can be categorized into decision-based and score-based attacks. Decision-based attacks~\cite{wang2025ttba} only have access to the predicted class label, whereas score-based attacks~\cite{rezagsba2025} can access the predicted probability distribution.

\begin{figure*}
    \centering
    \includegraphics[width=0.95\linewidth]{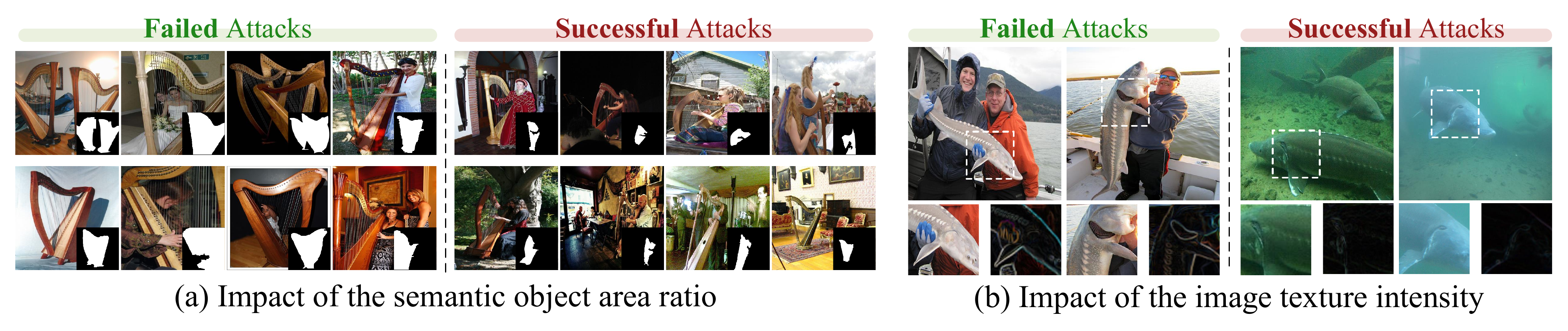}
    \caption{Illustrations of how semantic object area ratio and texture intensity affect the robustness of benign samples. DeCoWA is used as the attack with R50 as the surrogate model. The target model is D161 in (a) and is SwinT in (b).}
    \label{fig:illustration_samples}
\end{figure*}

\textit{Generation-based attacks} employ neural-network generators to craft adversarial examples. These generators are typically trained to either increase the output loss of a surrogate model~\cite{li2024ucg} or to shift its intermediate feature representations~\cite{li2025aim}.

This paper will evaluate the proposed measure under the aforementioned various advanced adversarial attacks.

\subsection{Model-Centric Robustness Analysis}
Conventional studies on adversarial robustness primarily focus on a model-centric perspective. 

\textit{Formal robustness certification}~\cite{zhang2024seev} uses formal methods to mathematically prove whether a neural network satisfies certain robustness properties. 

\textit{Robustness evaluation} relies on empirical methods to estimate a model's robustness. Representative approaches include evaluating the model's accuracy under selected adversarial attacks~\cite{xie2025towards} and estimating the geometric distance between a sample and the model's decision boundary~\cite{yan2024enhance}. 

\textit{Robust architecture design}~\cite{xu2025rethinking} explores how architectural choices affect model robustness, seeking structural variations that improve robustness even under standard training. 

\textit{Adversarial defenses} focus on arming models with carefully designed mechanisms to withstand adversarial threats. Representative strategies encompass adversarial training~\cite{liu2025comprehensive}, input purification~\cite{song2024mimicdiffusion}, randomized smoothing~\cite{chaouai2024universal}, and adversarial detection~\cite{li2024self}.

\subsection{Sample-Centric Robustness Analysis}
Unlike model-centric robustness analysis, recent studies reconsider from a sample-centric perspective. They aim to find measures for estimating image attackability. It is promising in active learning, adversarial training, and sample selection-based efficient attacks. It mainly includes two types: (1) evaluating the attackability of benign samples and (2) predicting the transferability of adversarial examples. ET~\cite{levy2024ranking} utilizes multiple surrogate models to approximate the expected transferability of adversarial examples. IAARS~\cite{raina2023identifying} measures the attackability of benign samples as the smallest perturbation required to change the prediction of the model proxy. They use a given dataset and model proxy to train a neural-network discriminator. Then, they utilize the discriminator to evaluate the attackability of unseen images on unseen models. ZGP~\cite{karunanayake2025quantifying} measures the attackability as the element ratio in the image gradients of the proxy's loss, whose absolute value is below a certain threshold.

However, ET, IAARS and ZGP are model-dependent. Moreover, the attackability features they extract cannot be directly linked to the image itself in a visually interpretable manner. To address these, this paper aims to develop a model-free and visually interpretable attackability measure.

\section{Methodology}

\subsection{Problem Definition}
Given an image set $D=\{(x,y)\in \mathcal{X}\times \mathcal{Y}\}$, where $\mathcal{X}$ and $\mathcal{Y}$ represent the image space and label space, respectively, for a target model $M$ and an attack $A$, we aim to find a model-free and visually interpretable measure $\phi^{*}(x)$ that satisfies
\begin{equation}
    \phi^{*} = \arg\max_{\phi}\frac{|\{(x,y)\in D_{\alpha}(\phi)|M(A(x))\neq y\}|}{\alpha|D|},
    \label{euqation:optimization_problem}
\end{equation}
in contrast to the conventional model-dependent one $\phi^{*}(x,M')$. Here, $|\cdot|$ indicates the cardinality of a set. $D_{\alpha}(\phi)$ is the top-$\alpha$ subset of attackable images from the set $D$, sorted according to the measure $\phi$. 
% It represents the subset of $D$ considered most vulnerable under the measure $\phi$. 
$0<\alpha<1$ is the sampling rate (SR). For convenience, we define the objective function in Equation~\ref{euqation:optimization_problem} as $\text{r}(D_{\alpha}(\phi), M, A)$.

However, optimization problem~\ref{euqation:optimization_problem} is intractable, and we thus relax it to a more solvable problem as below
\begin{equation}
\begin{aligned}
    \hat{\phi} &= \phi, \\
    \text{s.t.}~\text{r}(D_{\alpha}(\phi), M, A)-\text{r}(&D_{\alpha}(\phi_{\text{random}}), M, A)>\eta,
\end{aligned}
\end{equation}
where $\phi_{\text{random}}$ is the random evaluation strategy, and $\eta$ is the acceptable threshold for the attack success rate (ASR) gain.

\begin{figure*}
    \centering
    \includegraphics[width=0.95\linewidth]{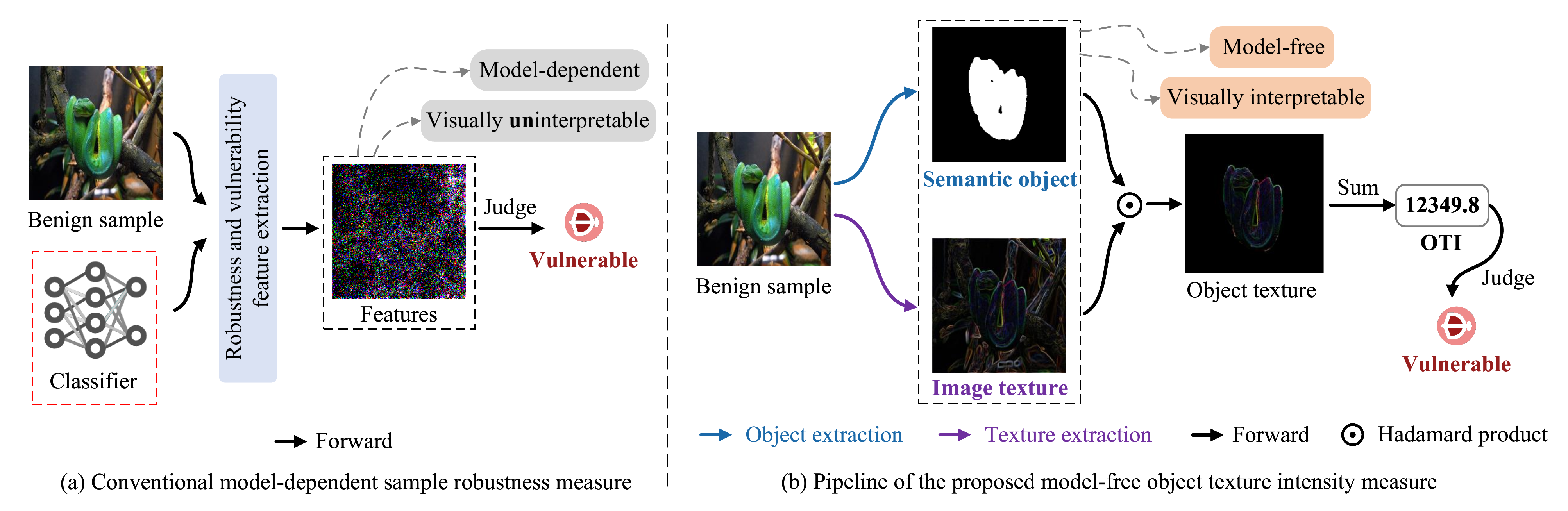}
    \caption{Comparison of overall pipelines between conventional methods and our proposed method.}
    \label{figure:pipeline}
\end{figure*}

\subsection{Motivation}
% To identify a model-free and visually interpretable measure of sample robustness, 
Comprehensive empirical studies across various attacks, tasks, datasets, and other configurations reveal that two types of images tend to be more attackable:
\begin{enumerate}[label=\arabic*)]
    \item Images with smaller semantic object area;
    \item Images with weaker textures and lower contrast.
\end{enumerate}
Illustrative examples based on the DeCoWA attack~\cite{lin2024boosting} are presented in Figure~\ref{fig:illustration_samples}. For the \texttt{harp} class, images that resist adversarial perturbations typically have larger harp objects than those that are successfully attacked. In the case of \texttt{sturgeon} class, low-contrast and weak-texture images taken underwater are generally more attackable compared to clearly captured images taken above water. Motivated by these two sample-intrinsic and human-perceivable trends, we propose the following measures.

\subsection{Object Texture Intensity}
We quantify the first trend using the \textbf{Object Area Ratio (OAR)}, which indicates the proportion of the image area occupied by task-relevant objects. OAR is formally defined as
\begin{equation}
\begin{aligned}
    \text{OAR}(x^{(C\times H\times W)}) = \frac{1}{C\times H\times W}||\text{object}(x)||_{1}.
    \label{equation: OAR}
\end{aligned}
\end{equation}
Here, the function $\text{object}(\cdot)$ is a binarization function that converts the input $x$ into a single-channel object segmentation map. It is worth noting that $\text{object}(\cdot)$ can be realized by semantic segmentation or salient object detection~\cite{liu2021visual} networks, gradient activation maps~\cite{selvaraju2017grad}, other coarse annotations, or precise manual labels, as long as it produces an approximately correct binarized object segmentation map.

For the second trend, we quantify it using \textbf{Image Texture Intensity (ITI)}, which indicates the sum of absolute texture values in an image. ITI is formulated as
\begin{equation}
\begin{aligned}
    \text{ITI}(x^{(C\times H\times W)}) =\frac{1}{C\times H\times W}||f*x||_{1},
    \label{equation: ITI}
\end{aligned}
\end{equation}
where $f$ denotes a texture extraction operator and $*$ denotes the convolution operation. While the choice of $f$ is flexible, we find that the Sobel operator~\cite{gonzales1987digital} suffices to effectively differentiate image attackability. To expose the intrinsic relationship between texture intensity and attackability, as well as improve computational efficiency, we adopt the Sobel operator as $f$ in this work.

% Furthermore, these two measures reflect different aspects of sample robustness. Therefore, we can integrate them to form a stronger sample-intrinsic robustness measure, termed Object Texture Intensity. We begin by formalizing the semantic object area ratio and image texture intensity.

OAR and ITI measure distinct aspects of an image. By integrating both, we can better leverage their complementary properties to construct a stronger measure, \textbf{Object Texture Intensity (OTI)}, which quantifies the texture intensity of the semantic object in an image. OTI is formulated as
\begin{equation}
\begin{aligned}
    \text{OTI}(x) =\frac{1}{C\times H\times W}||\text{object}(x)\odot(f*x)||_{1},
    \label{equation: OTI}
\end{aligned}
\end{equation}
where $\odot$ denotes the Hadamard product. Figure~\ref{figure:pipeline}(b) illustrates a complete pipeline of OTI.

\subsection{Explanation of the Measure}
In this section, we explain OTI's effectiveness from the perspectives of decision boundaries as well as the mid- and high-frequency characteristics of adversarial perturbations.

\begin{figure}
    \centering
    \includegraphics[width=0.96\linewidth]{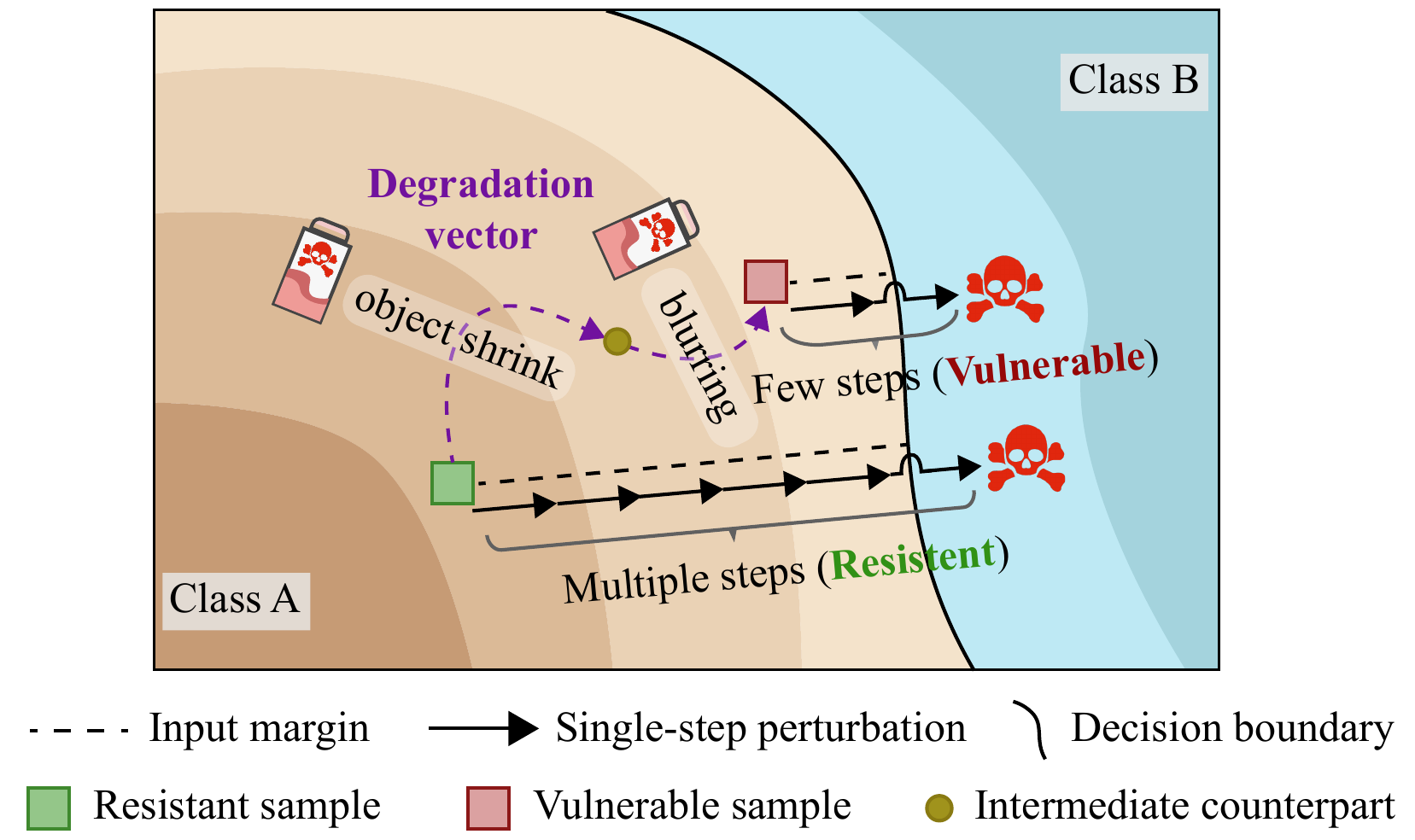}
    \caption{Illustration explaining the effectiveness of OTI based on model decision boundaries.}
    \label{fig:explanation}
\end{figure}

\begin{figure*}
    \centering
    \includegraphics[width=0.98\linewidth]{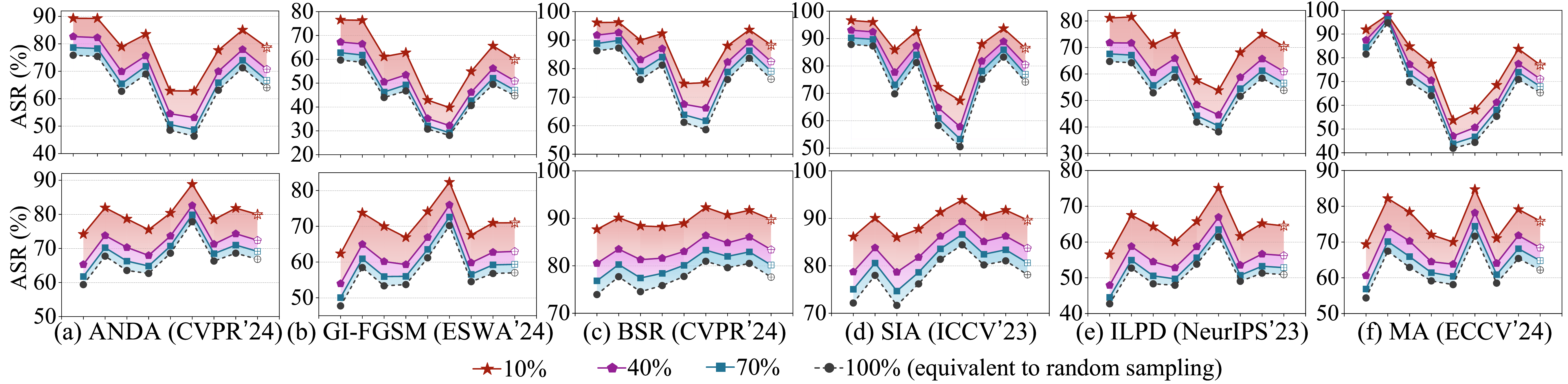}
    \caption{ASRs under varying sampling rates $\alpha$ in cross-model untargeted attacks. The surrogate models are ResNet-50 (top row) and ViT-B/16 (bottom row). The x-axis represents the target models, with the last one indicating the average.}
    \label{figure:experiement_single_surrogate_model_untargeted}
\end{figure*}

\textit{Explanation based on decision boundaries.}
In the model-centric robustness analysis, it is widely believed that a primary source of model vulnerability is the small input margin~\cite{elsayed2018large, yan2024enhance}, where the decision boundary is too close to the input sample, making the model susceptible to even minor perturbations. Shifting to a sample-centric perspective, as illustrated in Figure~\ref{fig:explanation}, attackable samples can be pushed across the decision boundary with only a few steps of perturbation, whereas robust samples require significantly more steps. Thus, the image attackability can essentially be characterized by the image's distance to the decision boundary. Going a step further, a robust image can be converted into a vulnerable counterpart by a degradation vector. Applying certain transformations to a robust image can bring it closer to the decision boundary, thereby increasing its attackability. For example, downscaling the semantic object reduces the amount of discriminative information and lowers the signal-to-noise ratio~\cite{pawlowski2019needles}, bringing the sample closer to the decision boundary. Likewise, image blurring leads to the loss of high-frequency components~\cite{zhang2024fourier}, which also moves the sample closer to the decision boundary. Therefore, OTI essentially estimates a relative degradation vector based on the object size and image texture, which in turn approximates the distance to the decision boundary. A lower OTI indicates a larger degradation vector, meaning the sample lies closer to the decision boundary and is thus more attackable.

\textit{Explanation based on mid- and high-frequency characteristics of adversarial perturbations.} The mid- and high-frequency components are generally considered to contain higher energy in adversarial perturbations~\cite{wang2020high}, especially the recent study~\cite{chen2025intriguing} that highlights a positive correlation between high-frequency components and attack performance. OTI quantifies the intensity of semantically relevant mid- and high-frequency information. A lower OTI suggests that the image contains fewer mid- and high-frequency components, making it less capable of resisting the high-energy mid- and high-frequency components of adversarial perturbations, and thus more attackable.

In summary, from the perspective of decision boundaries, OTI characterizes the magnitude of the degradation vector. From the frequency-domain perspective, OTI reflects the amount of semantically relevant mid- and high-frequency components. An image's attackability to adversarial perturbations can thus be inferred from its semantic object size and the intensity of its texture. In this way, we establish a visual connection between an image and its attackability.

% To conclude, OTI offers both simplicity and visual interpretability, which facilitates a clearer understanding of its underlying principles. It also provides a practical visual basis for estimating the robustness of a given sample. Furthermore, since OTI relies solely on the sample itself rather than any model-specific information, it is inherently model-free. This property enables OTI to generalize well in scenarios involving transfer-based attacks.

\section{Experiments and Results}
% In this section, we empirically demonstrate the effectiveness of OTI in evaluating image attackability. 
% In this section, we demonstrate the utility of OTI in estimating sample robustness through extensive experiments.
\subsection{Setup}
\subsubsection{Datasets.}
To evaluate our OTI on sets of different scales, we adopt the large-scale ImageNet Validation dataset~\cite{deng2009imagenet} and a small-scale subset of $1,000$ images for natural image classification. Additionally, to assess OTI across different tasks and domains, we use Kvasir-SEG~\cite{jha2019kvasir} for gastrointestinal polyp segmentation.
% These three datasets allow us to assess OTI under varying tasks, dataset sizes, and data domains. 

\subsubsection{Evaluation benchmarks.} Diverse advanced attacks from various strategies are adopted as benchmarks, including gradient-optimization methods ANDA~\cite{fang2024strong} and GI-FGSM~\cite{wang2024boostinggi}, transformation-based methods SIA~\cite{wang2023structure} and BSR~\cite{wang2024boosting}, objective-optimization method ILPD~\cite{li2023improving}, model-related method MA~\cite{ma2024improving}, ensemble-based methods Ens~\cite{liu2016delving}, AdaEA~\cite{chen2023adaptive}, SMR~\cite{tang2024ensemble}, and CWA~\cite{chen2024rethinking}, query-based methods ADBA~\cite{wang2025adba}, TtBA~\cite{wang2025ttba}, and methods specifically designed for targeted attacks SU~\cite{wei2023enhancing}, Logit Margin~\cite{weng2023logit}, CFM~\cite{byun2023introducing} and FFT~\cite{zeng2024enhancing}.

\subsubsection{Models.} Single surrogate models used are ViT-B/16 and R50. Ensemble surrogate models used include E$_{1}$:\{R18, R50, R101\}, E$_{2}$:\{MV2, IncV3, BeiT-B/16\}, and E$_{3}$:\{ViT-S/16, ViT-S/32, ViT-B/32\}. Undefended target models include R152, D161, IncRes-V2, ConvNeXt-B, Swin-B, BeiT-B/16, XCiT-S, and Poolformer. The defended adversarially trained models include ConvNeXt-B + ConvStem~\cite{singh2023revisiting}, ViT-B + ConvStem~\cite{singh2023revisiting}, ConvNeXt-B~\cite{liu2025comprehensive}, SwinT-B~\cite{liu2025comprehensive}, RaWideR101-2~\cite{peng2023robust}, and ConvNeXtV2-L + Swin-L~\cite{bai2024mixednuts}. The target models used for segmentation include: U-Net, UNet-CCT, UNet-URPC, MWCNN, ATTU-Net, and ResU-Net. The experimental results are organized by default in the same order as presented here.

\begin{figure*}
    \centering
    \includegraphics[width=0.92\linewidth]{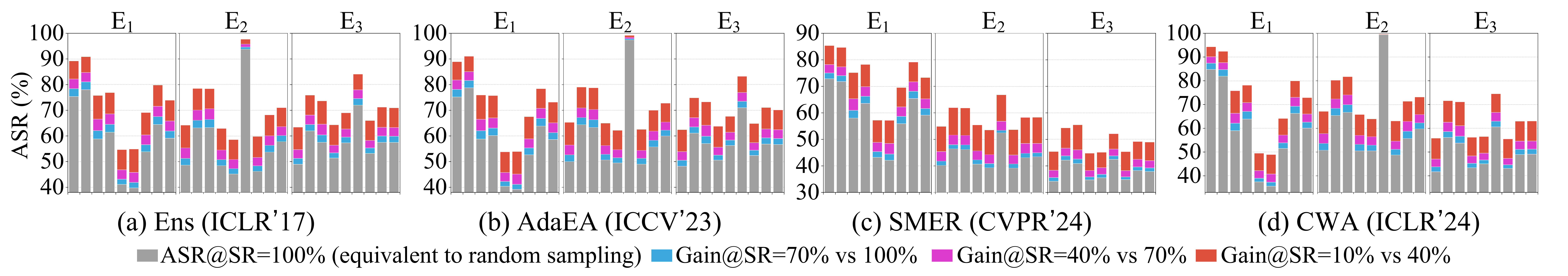}
    \caption{ASRs under varying sampling rates $\alpha$ in ensemble-based attacks. The x-axis represents the target models, with the last one indicating the average.}
    \label{figure:experiement_ensemble_based_untargeted}
\end{figure*}

\subsubsection{Metrics.} For classification tasks, we use attack success rate (ASR) as the metric, defined as the proportion of successfully attacked samples over the total number of samples. For segmentation tasks, we adopt the precision, F1-score, and IoU as metrics. Given a sampling rate $\alpha$ for vulnerable images, a higher ASR or lower precision, F1-score, and IoU suggest that OTI achieves more accurate identification and thus demonstrates a stronger capability in evaluating image attackability. Experiments are conducted on NVIDIA A100 Tensor Core GPU, and the results reported are based on a single run.

\subsubsection{Details of Implementation.} For the ImageNet validation dataset, where the semantic and salient objects largely overlap, we use the saliency detector VST~\cite{liu2021visual} as the object segmenter. For Kvasir-SEG, we simply use the manually annotated groundtruth as the segmentation map. 
% The experiments are conducted on NVIDIA A100 Tensor Core GPU, and the results reported are based on a single run.

\begin{figure}
    \centering
    \includegraphics[width=0.92\linewidth]{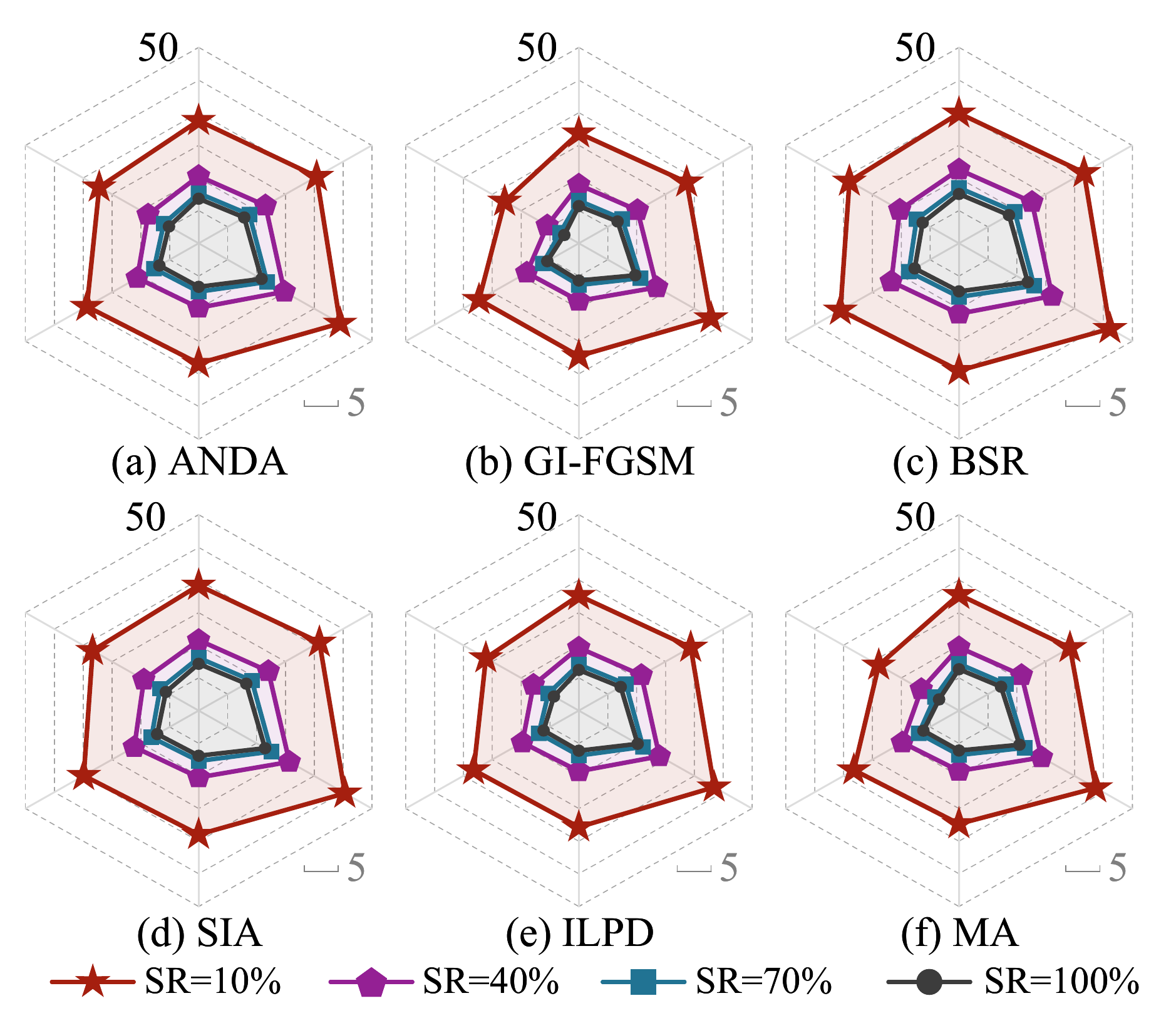}
    \caption{ASRs under varying sampling rates $\alpha$ in attacking adversarially trained models. The first surrogate model is at the top of the hexagon, with others arrange clockwise.}
    \label{experiment:attack_adv_trained}
\end{figure}

\begin{figure}[t]
    \centering
    \includegraphics[width=0.94\linewidth]{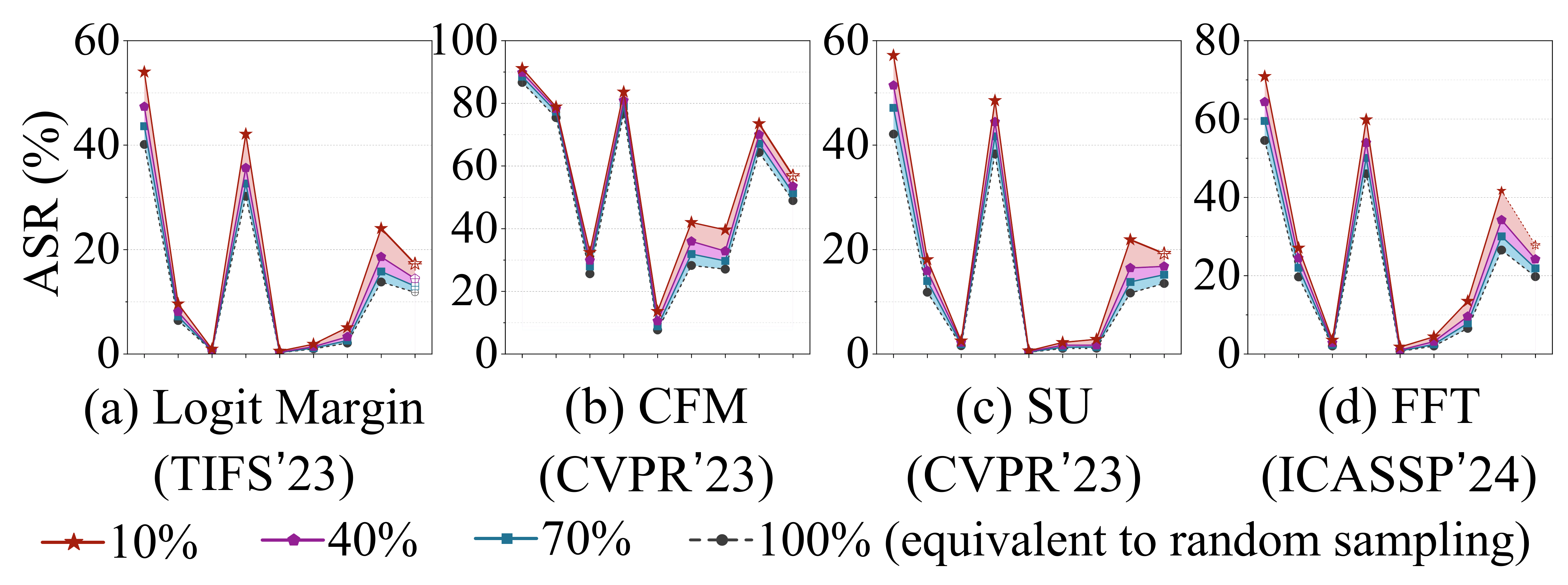}
    \caption{ASRs under varying sampling rates $\alpha$ in targeted attacks. The x-axis represents the target models, with the last one indicating the average.}
    \label{experiment:single_surrogate_model_targeted}
\end{figure}

\begin{figure}
    \centering
    \includegraphics[width=0.94\linewidth]{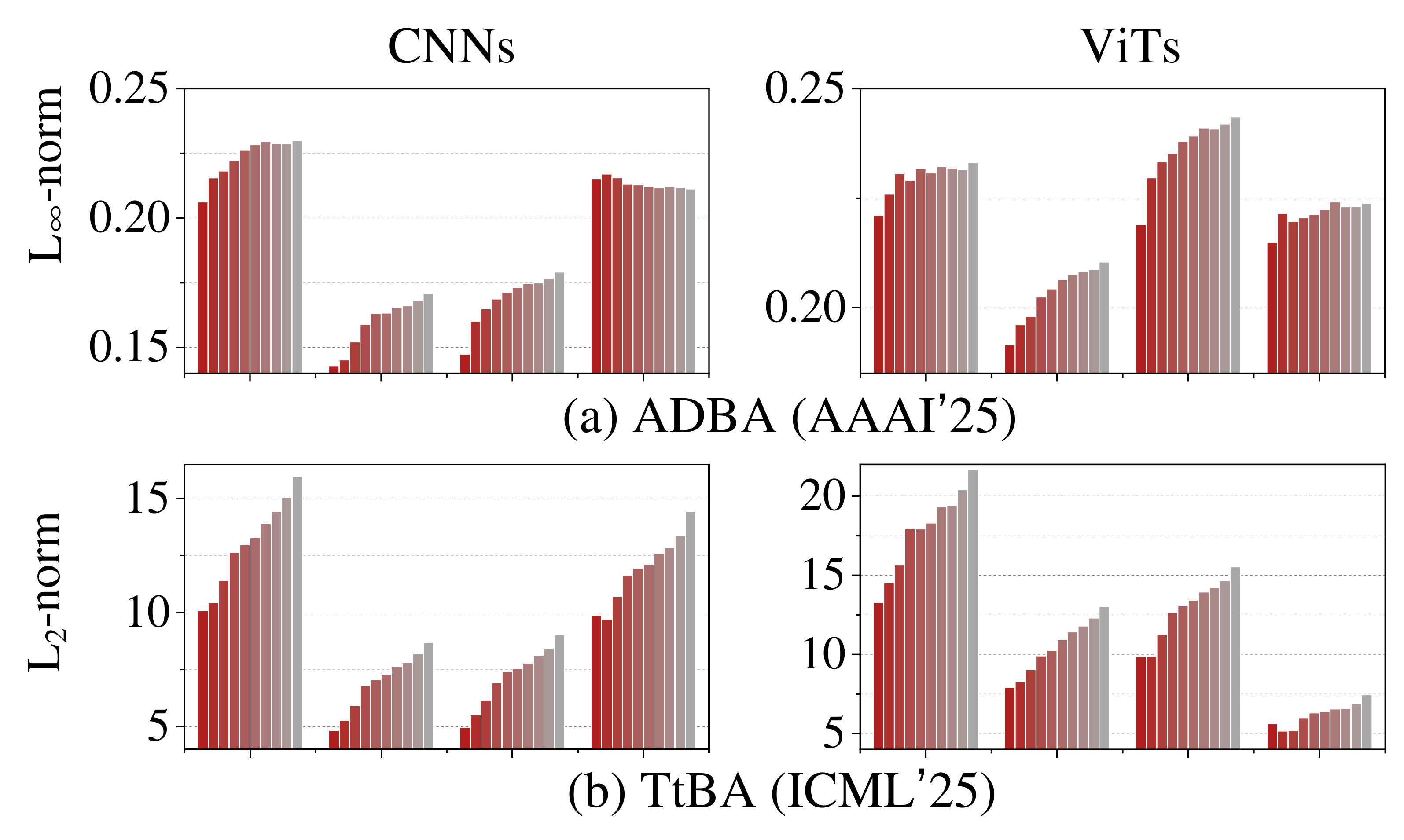}
    \caption{The query-based perturbation norm required for samples at different sampling rates. Each group corresponds to a target model. With each group, the 10 bars represent sampling rates ranging from $10\%$ to $100\%$ in $10\%$ intervals.}
    \label{experiment:query_based_attacks}
\end{figure}

\subsection{Evaluation of Robustness Discrimination}
% \subsubsection{Under white-box single-surrogate untargeted attacks.}
\subsubsection{Under single-surrogate untargeted attacks.}
In this experiment, we evaluate OTI under single-surrogate untargeted attacks. Specifically, we rank the $50K$ benign images from the ImageNet Validation dataset by OTI, and select the top-$\alpha$ most attackable images. These images are then used to craft adversarial examples on the surrogate model, which are transferred to attack black-box target models. We report ASRs across varying $\alpha$ values. For comparison, we also report ASRs of the random sampling strategy (equivalent to OTI-based sampling of $\alpha=100\%$). A larger ASR gap between OTI-based and random sampling at different sampling rates indicates better performance of OTI. We select $6$ advanced benchmarks. The surrogate model and target models are as described in the Setup. The perturbation budget is $L_{\infty}=10/255$. The results are illustrated in Figure~\ref{figure:experiement_single_surrogate_model_untargeted}. Experimental results show that images selected at lower sampling rates achieve higher ASRs. At $\alpha=10\%$, the average ASR improves by $13.39\%$ compared to random sampling, demonstrating OTI's effectiveness.

\begin{figure*}
    \centering
    \includegraphics[width=0.92\linewidth]{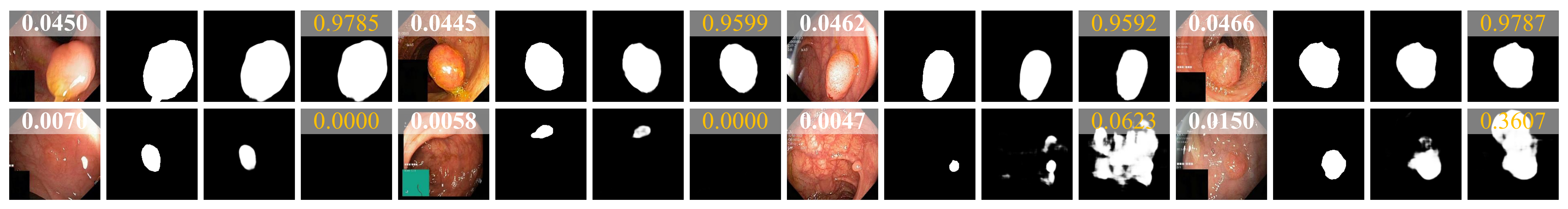}
    \caption{Visualization of attacks on Kvasir-SEG. The first row shows high-OTI samples, and the second row shows low-OTI samples. White text indicates OTI values,  and yellow text indicates F1-scores. Each group shows: benign sample, groundtruth, benign prediction and attacked prediction. The surrogate model is URPC and target model is U-Net.}
    \label{figure:kvasir_visualization}
\end{figure*}

\begin{figure*}
    \centering
    \includegraphics[width=0.92\linewidth]{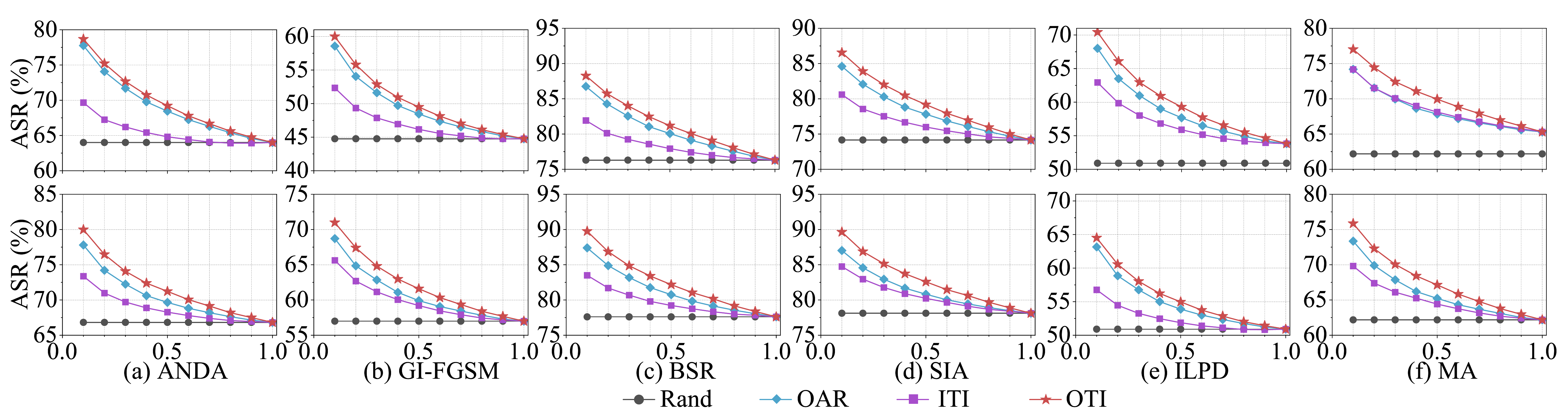}
    \caption{Ablation study of OTI. The surrogate models are ResNet-50 (top row) and ViT-B/16 (bottom row). The x-axis indicates the samping rate, and the y-axis shows the average ASR across $8$ targets.}
    \label{figure:ablation_OAR_ITI_OTI}
\end{figure*}

\subsubsection{Under single-surrogate targeted attacks.}
This experiment evaluates OTI to identify image attackability to targeted attacks, following a process similar to that used for untargeted attacks. The surrogate model is R50 and the perturbation budget is $L_{\infty}=16/255$. The results shown in Figure~\ref{experiment:single_surrogate_model_targeted} indicate that, despite the increased difficulty of targeted attacks, subsets selected based on OTI achieve higher ASRs. On average, at $\alpha=10\%$, OTI-based sampling improves the ASR by $6.79\%$ compared to random sampling.

\subsubsection{Under ensemble-based untergeted attacks.}
This experiment investigates OTI to evaluate attackability to ensemble-based attacks. Four advanced ensemble-based benchmarks and three ensemble surrogate models are as described in Setup. The perturbation budget is $L_{\infty}=10/255$. Experimental results presented in Figure~\ref{figure:experiement_ensemble_based_untargeted} indicate that OTI is also effective in evaluating image attackability to ensemble-based attacks, achieving an average improvement of $13.55\%$ over random sampling at $\alpha=10\%$.

\subsubsection{On attacking adversarially trained models.} In the above experiments, target models are all undefended. To test OTI's performance on defended models, we use R50 as the surrogate model and evaluate on six advanced adversarially trained models as described in Setup. Results in Figure~\ref{experiment:attack_adv_trained} confirm that OTI reliably assesses image attackability even with defenses. On average, OTI-based sampling achieves a $12.24\%$ higher ASR than random sampling at $\alpha=10\%$.

\subsubsection{Query optimization in query-based attacks.}
Since OTI effectively measures image attackability, more attackable images require less perturbation. Intuitively, OTI-selected vulnerable images need smaller perturbations. We evaluate on two SOTA query-based methods with $L_{\infty}$ and $L_{2}$ norms respectively. The dataset used is the subset of ImageNet Validation. Results in Figure~\ref{experiment:query_based_attacks} confirm our thoughts.

\subsubsection{On attacking gastrointestinal polyp segmentation.}
\begin{figure}
    \centering
    \includegraphics[width=0.94\linewidth]{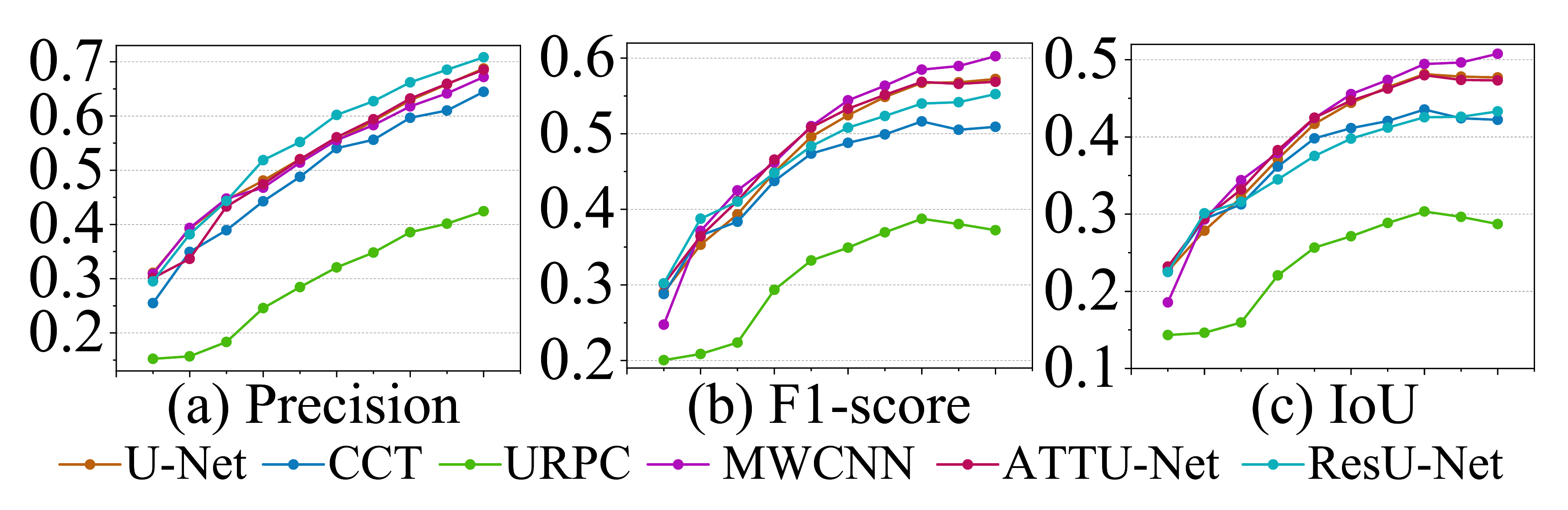}
    \caption{Metrics of MI-FGSM based Kvasir-SEG adversarial examples at different sampling rates. The x-axis represents sampling rates from $10\%$ to $100\%$ in $10\%$ intervals.}
    \label{experiment:attack_kvasir}
\end{figure}

Extensive experiments have shown OTI's effectiveness in classification tasks. We further investigate whether OTI remains effective in segmentation tasks with non-natural images, and whether it performs well with a smaller dataset scale compared to previous datasets of $50K$ and $1K$ images. To this end, we test on the $200$-image Kvasir-SEG dataset using the MI-FGSM attack with a perturbation budget of $L_{\infty}=4/255$. The surrogate model is U-Net. The target models are shown in Setup. Visualization results are shown in Figure~\ref{figure:kvasir_visualization}, with detailed data in Figure~\ref{experiment:attack_kvasir}. Both confirm that OTI remains effective in non-natural image segmentation tasks.

\subsection{Ablation Study}
This experiment aims to ablate OTI to explore the relationship among OAR, ITI, and OTI. We use untargeted single-surrogate attacks as the benchmark. The results are shown in Figure~\ref{experiment:query_based_attacks}. It can be observed that in all cases, OTI outperforms OAR and ITI, indicating that OAR and ITI are complementary robustness features. Additionally, OAR generally outperforms ITI because OAR focuses on the quantity of semantically relevant elements, while ITI considers global and possibly semantic irrelevant texture information. OAR, ITI, and OTI all significantly outperform random strategy, demonstrating their effectiveness.

\section{Conclusions and Limitations}
This paper proposes OTI, a model-free and visually interpretable measure for evaluating the benign image attackability to adversarial perturbations. Our OTI aligns with human intuition and performs well across various tasks, domains, and data sizes. However, it is limited to image data, making the development of similar measures for audio, text, and other modalities a valuable future direction.

\section*{Acknowledgments}
This work was supported in part by the Science and Technology Development Fund, Macau SAR, under Grant 0193/2023/RIA3 and 0079/2025/AFJ, and the University of Macau under Grant MYRG-GRG2024-00065-FST-UMDF.

\bibliography{aaai2026}

\end{document}